\documentclass[11pt,a4paper]{article}
\usepackage{times,latexsym}
\usepackage{url}
\usepackage[T1]{fontenc}
\usepackage[acceptedWithA]{tacl2021v1}

\usepackage{amsthm}
\usepackage{amsmath,amsfonts,bm}
\makeatletter
\newtheorem*{rep@theorem}{\rep@title}
\newcommand{\newreptheorem}[2]{%
\newenvironment{rep#1}[1]{%
 \def\rep@title{#2 \ref{##1}}%
 \begin{rep@theorem}}%
 {\end{rep@theorem}}}
\makeatother

\newreptheorem{theorem}{Theorem}

\newcommand{\RNum}[1]{\uppercase\expandafter{\romannumeral #1\relax}}

\usepackage{mathtools}

\newcommand{\cut}[1]{}

\newcommand{\removelatexerror}{\let\@latex@error\@gobble}

\def\eqref#1{Eq.~\ref{#1}}

\def\1{\bm{1}}

\DeclareMathAlphabet{\mathsfit}{\encodingdefault}{\sfdefault}{m}{sl}
\SetMathAlphabet{\mathsfit}{bold}{\encodingdefault}{\sfdefault}{bx}{n}

\usepackage{times}
\usepackage{xspace,mfirstuc,tabulary}
\usepackage{latexsym}
\usepackage{linguex}
\usepackage{graphicx}
\usepackage{booktabs}
\usepackage{multirow, makecell}
\usepackage{color, colortbl}

\usepackage{etoolbox}
\usepackage{numprint}
\patchcmd{\quote}{\rightmargin}{\leftmargin 1em \rightmargin}{}{}
\usepackage{multirow}
\usepackage{microtype}
\usepackage{tikz}
\usepackage{amssymb}
\usepackage{mdframed}
\usepackage{cuted}
\usepackage{subfigure}

\definecolor{gray}{rgb}{.95, .95, .95}
\definecolor{highlight}{rgb}{1.0, 1.0, 0.89}
\definecolor{aquamarine}{rgb}{0.5, 1.0, 0.83}
\definecolor{asparagus}{rgb}{0.53, 0.66, 0.42}
\definecolor{emerald}{rgb}{0.31, 0.78, 0.47}
\definecolor{gainsboro}{rgb}{0.86, 0.86, 0.86}
\definecolor{gold(web)(golden)}{rgb}{1.0, 0.84, 0.0}

\theoremstyle{remark}

\newcommand{\CTRL}{\textsc{CTRL-Dialog}}

\newcommand{\BEGIN}{\textsc{Begin}}
\newcommand{\begindata}{\BEGIN}

\newcolumntype{L}[1]{>{\raggedright\let\newline\\\arraybackslash\hspace{0pt}}p{#1}}

\title{Evaluating Attribution in Dialogue Systems: The BEGIN Benchmark
}
 \author{
  Nouha Dziri\hspace*{.4mm}\Thanks{Equal Contribution.}\hspace*{.4mm}\ \Thanks{Work done while at Google Research.}\hspace*{.2mm} $^{\diamondsuit}$ \quad Hannah Rashkin$^{*\hspace*{.1mm}\S}$ \quad Tal Linzen$^{\S\hspace*{.15mm}\clubsuit}$ \quad David Reitter$^\S$
   \\
  $^\diamondsuit$University of Alberta, Canada \quad $^\S$Google Research, USA \quad $^\clubsuit$New York University, USA  \\
  \texttt{dziri@cs.ualberta.ca} \quad
  { \tt \{hrashkin,linzen,reitter\}@google.com} 
 }

\date{}

\begin{document}

\setlength{\textfloatsep}{10pt plus 2.0pt minus 4.0pt}

\setlength{\Extopsep}{0.2\baselineskip}
\setlength{\Exredux}{0\baselineskip}
\setlength{\Exlabelwidth}{1em}%

\maketitle
\begin{abstract}
Knowledge-grounded dialogue systems powered by large language models often generate responses that, while fluent, are not \textit{attributable} to a relevant source of information. Progress towards models that do not exhibit this issue requires evaluation metrics that can quantify its prevalence. To this end, we introduce the Benchmark for Evaluation of Grounded INteraction (\begindata), comprised of 12k dialogue turns generated by neural dialogue systems trained on three knowledge-grounded dialogue corpora. We collect human annotations assessing the extent to which the models' responses can be attributed to the given background information. We then use \begindata\ to analyze eight evaluation metrics. We find that these metrics rely on spurious correlations, do not reliably distinguish attributable abstractive responses from unattributable ones, and perform substantially worse when the knowledge source is longer. Our findings underscore the need for more sophisticated and robust evaluation metrics for knowledge-grounded dialogue. We make \begindata \ publicly available at \url{https://github.com/google/BEGIN-dataset}.

\end{abstract}

\section{Introduction}

\begin{figure}[ht]
\centering
\includegraphics[width=\columnwidth]{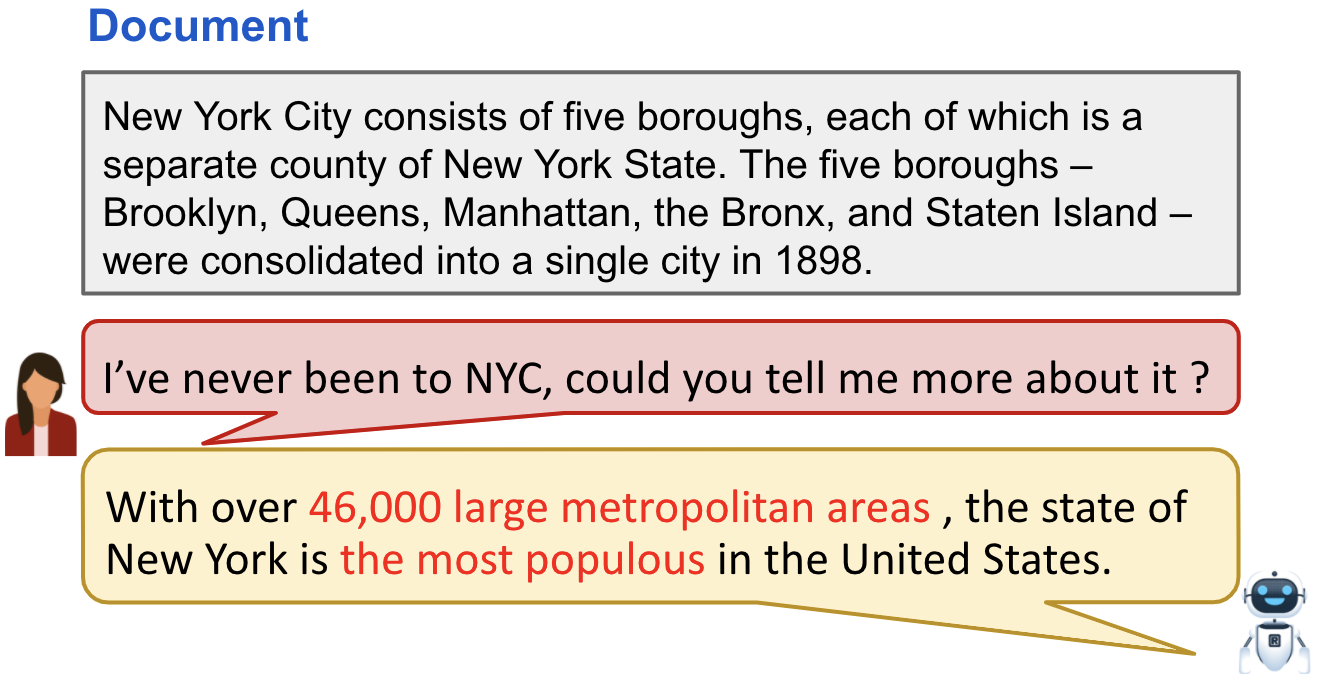}
\caption{An example of a response generated by the GPT2 language model fine-tuned on the Wizard of Wikipedia dataset \cite{dinan2018wizard}. The phrases in red are ``hallucinations'' unsupported by the background document. \label{fig:hallucination}}
\end{figure}

Neural language models \citep[][\textit{inter alia}]{bengio2000neural, vaswani2017attention, radford2019language} often form the backbone of open-ended dialogue systems \cite{wolf2019transfertransfo, zhang2019dialogpt, roller-etal-2021-recipes, adiwardana2020towards}. Utterances sampled from such language models sound natural, as reflected in these systems' high scores in human evaluations focused on measures such as ``engagingness'' or ``human-likeness'' \cite{see-etal-2019-makes}. While fluent, however, the responses generated by these systems often contain statements that are not supported by the evidence available to the system; such statements are sometimes referred to informally as ``hallucinations'' (\citealt{tian2020sticking, maynez-etal-2020-faithfulness, dziri-etal-2021-neural, shuster-etal-2021-retrieval-augmentation}; see Figure~\ref{fig:hallucination} for an example). This issue is particularly salient for knowledge-grounded dialogue systems, which are expected to interact with a user in an open-ended fashion while conveying information that is \textit{attributable} to external identifiable sources.
In this work, we develop a benchmark that can be used to assess attribution in knowledge-based dialog systems; following \citet{rashkin2021measuring}, we define
an attributable response\footnote{Attribution is sometimes referred to as faithfulness \citep[\textit{inter alia}]{cao-etal-2018,durmus-etal-2020-feqa}.} as one connected to textual evidence that supports the entirety of the response.

A number of modelling approaches have recently been proposed to increase attribution in knowledge-grounded dialog systems \cite{ rashkin-etal-2021-increasing, shuster-etal-2021-retrieval-augmentation, dziri-etal-2021-neural, dziri2022faithdial}. Progress in this area crucially relies on metrics that can measure the attribution of the text generated by the system; and indeed, recent work  has developed automated metrics with relatively high correlations with human annotations, potentially paving the way for alternatives to expensive human evaluations \cite{honovich-etal-2021-q2, dziri-etal-2021-neural, dziri2022faithdial}. Yet our understanding of these recently proposed metrics, as well as more established ones, remains limited, for two reasons.
First, comparisons between automated metrics and human judgments rely on small-scale datasets with a few hundred examples. This results in high variance in our estimate of the correlation coefficient and a limited ability to measure performance on infrequent example types \cite{gehrmann2021gem}.
Second, the correlation with human scores does not sufficiently determine the efficacy and robustness of automatic metrics produced by neural networks: such learned metrics---like other properties learned by neural networks---can be susceptible to spurious correlations that fail to generalize to more challenging cases.
To address these limitations, we introduce a large-scale resource, the Benchmark for Evaluation of Grounded INteraction (\begindata),  for meta-evaluation of metrics designed to evaluate grounded dialogue. In other words, the goal of this benchmark is to determine to what extent current evaluation metrics fulfill their purpose.

We define a taxonomy dividing knowledge-grounded dialogue responses into three broad categories---\textit{fully attributable}, \textit{not fully attributable}, and \textit{generic}---and ask humans to classify a large set of utterances produced by dialogue systems with this taxonomy. The motivation for the \textit{generic} category we introduce---which is assigned to utterances such as ``\textit{Sorry, I'm not sure about this topic}''---is the intuition that evaluation metrics should not treat the basic elements of a natural-sounding conversation, such as backchanneling or acknowledgment \cite{grice1989studies, stiles1992describing, bunt-etal-2020-iso}, as equally undesirable as a misleading unattributable statement. In real-world scenarios, it is  preferable for a model to acknowledge its ignorance instead of producing hallucinated content which may lead to the spread of disinformation.
 
Using this taxonomy, we then collect high-quality human annotations for 12k examples generated by four language-model-based dialogue systems, each trained on three different knowledge-grounded dialogue corpora. Examples of machine-generated responses along with labels are presented in Table \ref{tab:benchmark:wow_cmu_begin_dist}. 
We use this benchmark to evaluate multiple existing automatic metrics including word-overlap measures, embedding-based measures, metrics based on Question Answering (QA) systems, and ones based on Natural Language Inference (NLI). We also propose a classifier  trained on an adversarially generated dataset we create. We find that all metrics inadequately measure attribution and all rely on spurious correlations to a large extent. In particular, the metrics tend to misidentify cases that are attributable but highly abstractive, as well as cases that are not fully attributable but use multiple words from the evidence document (i.e., unattributable but extractive). We also find that the metrics fail to measure attribution under distribution shift, scoring responses that pertain to relatively long knowledge sources the lowest. These results are in line with the robustness issues reported for other natural language generation
metrics, despite the high  correlation of those metrics with human judgments \cite{durmus-etal-2022-spurious, gehrmann2021gem, gabriel-etal-2021-go, yeh-etal-2021-comprehensive}.
We hope that \BEGIN{} will facilitate progress toward more robust metrics for grounded dialogue response generation.

\begin{table*}[ht]
\footnotesize
    \centering
    \begin{tabular}{c L{2cm} L{2cm} l L{7cm}}
    \toprule
    & \textbf{Category} & \textbf{Frequency} & \textbf{Example} \\
    \midrule
  \multirow{10}{*}{\rotatebox[origin=c]{90}{ \texttt{\textbf{WoW}}}} &  Fully Attributable  & 23.5\% &
        Knowledge: & \textbf{Raised in Compton, California, Lamar embarked on his musical career as a teenager} under the stage name k-dot, releasing a mix-tape that garnered local attention and led to his signing with indie record label top dawg entertainment.\\
        &&
        & History: & oh, yeah. I believe I've heard about that. The album has to be good then for sure. Do you know anything else about him or rap? \\
        &&&
        \textbf{T5 Response}: & Oh yes, I know that  \textcolor{teal}{the singer was raised in Compton, California and started on his musical career as a teenager}. 
        \\ \midrule
\multirow{10}{*}{\rotatebox[origin=c]{90}{ \texttt{\textbf{CMU-DoG}}}} &  Generic & 6.4\%  &
        Knowledge: & Throughout the film, the narrative is intercut with scenes from depositions taken in the Winklevoss twins' and Saverin's respective lawsuits against Zuckerberg and Facebook. ... After everyone leaves, Zuckerberg is shown sending a friend request to Albright on Facebook and then refreshing the webpage every few seconds as he waits for her response... \\
        &&& History: & The ending seems cools. Like after all of that happens he just goes to hang out with his kids.\\
        &&& \textbf{GPT2 Response}: & \textcolor{blue}{What else do you know about Zuckerberg?}
                            \\ \midrule

     \multirow{12}{*}{\rotatebox[origin=c]{90}{ \texttt{\textbf{TopicalChat}}}}  & Not Attributable & 70.1\% & 
        Knowledge: & Television (TV) is a telecommunication medium used for transmitting moving images in monochrome (black and white) .... Ringo Starr, George Carlin, and Alec Baldwin have all narrated Thomas the Tank Engine for at least 52 episodes each. There are over 750 episodes of "Pokémon": That's about 50 a year for 16 years straight.\\
        &&& History: & no, I haven't watched Pokemon. My sons use to play with when they were little. Now, they use their phones looking for Pokemon in the neighborhood. It is funny.\\
        &&& \textbf{DoHA Response}: & Pokemon \textcolor{red}{appeared in TV in 1952 and it was created by Ringo Starr and directed by George Carlin. In total, there were 100 episodes}.\\

        \bottomrule
    \end{tabular}
    \caption{\small Examples of each of the three categories of responses included in \begindata{}. For each category, we provide an example drawn from one of the four models trained on one of the three corpora (of course, all 12 models generated all three types of responses). The dialogue corpus used to train the model and generate the response is listed vertically. Text highlighted in green indicates information that is attributable to the knowledge; text in blue does not convey any information; and text in red is hallucinated and cannot be attributed to the knowledge.}
    \label{tab:benchmark:wow_cmu_begin_dist}
\end{table*}

\section{Task, Datasets and Models}
\label{sec:setup}

In {knowledge-grounded response generation}, the system is given a dialogue history
$\mathcal{H}=(u_1, \dots, u_{n-1})$, and knowledge $\mathcal{K}_n = (k_1, \dots, k_{j})$ at turn $n$, and is expected to generate a response $\bar{u}_{n}$ that is coherent with $\mathcal{H}$ and attributable to a non-empty subset $M_n \subset \mathcal{K}_n$.  Similar to the conversational QA task \cite{choi-etal-2018-quac, reddy-etal-2019-coqa}, the system is expected to use knowledge to respond to the user query. However, since the previous utterance may be an open-ended statement rather than a direct question (see the second and third examples in Table~\ref{tab:benchmark:wow_cmu_begin_dist}), there is a wider range of possible types of informative replies compared to the conversational QA task.

\textsc{Begin} consists of responses generated by language-model-based systems trained to perform this task. This section describes the models we train on this task and the corpora we use to train them.

\subsection{Dialogue Datasets}

For all three datasets, we use the training portion to train the model, the development set to tune hyperparameters, and the test set to generate the responses that are then annotated and included in the final \begindata{} benchmark.

\paragraph{Wizard of Wikipedia (WoW)} \textsc{WoW} dialogue \cite{dinan2018wizard} takes place between a Wizard and an Apprentice. The Wizard is tasked with providing information about a particular topic and the Apprentice, in turn, is expected to seek more information.
At each turn of the conversation, the Wizard is presented with passages from Wikipedia and chooses a span from the document---typically one or two sentences---that serves as evidence supporting their response.
We omitted examples where the Wizard did not explicitly select a passage as evidence for the response or where there was no dialogue history.  We also use the ``unseen'' topic portion of the test data. Overall, we used 82722 training examples, 8800 development examples, and 3902 test examples.

\paragraph{CMU-DoG} The CMU-DoG dataset \cite{zhou-etal-2018-dataset} consists of conversations about films. Each response is expected to be grounded in a section from Wikipedia. Workers can have either asymmetric or symmetric roles. In the asymmetric setting, one worker is asked to persuade the interlocutor to watch the movie using arguments from the document where only the persuader has access to the document. In the symmetric role, workers discuss together the content of the document. In total, there are 78136, 13800 and 13796 grounded responses (training/dev/test).
  
\paragraph{TopicalChat} TopicalChat \cite{Gopalakrishnan2019} consists of dialogues about a variety of topics. Workers are provided relevant facts from Reddit, Wikipedia and news articles. Analogous to \textsc{CMU-DoG}, the data collection protocol consists of two scenarios. In the symmetric scenario, workers have access to the same knowledge source; in the asymmetric scenario, they have access to different sources. They are asked to use the information from the documents to chat knowledgeably about the topic. In total, the dataset has 134572, 8790 and 8081 grounded responses (training/dev/test).

\subsection{Dialogue Models} We consider the outputs of four different dialogue systems; by selecting a relatively wide range of systems, we hope to encounter a range of attribution errors. Two of the systems are based on plain language models, GPT2-base \cite{radford2019language} and T5-base \cite{raffel2020exploring}. %
 The remaining two systems, DoHA \cite{prabhumoye-etal-2021-focused} and \CTRL{} \cite{rashkin-etal-2021-increasing}, are specifically designed as knowledge-grounded dialogue systems. 
DoHA augments a BART-based conversational model \cite{lewis2020bart} with a two-view attention mechanism that handles  the encoded document and the dilaogue history separately during generation. \CTRL{} augments T5-base with control tokens~\cite{keskar2019ctrl} that guide the generation towards less subjective and more grounded content.  We trained these models to generate responses based on a concatenation of two inputs: an evidence span (the knowledge snippet) and the dialogue history (we only use the previous turn $u_{n-1}$).

\section{Annotations}

We next describe the human annotations we collected for the utterances generated by the models described in Section~\ref{sec:setup}.

\subsection{Taxonomy of Response Types}
We classify responses into three broad categories:

\paragraph{Fully Attributable}  These are responses that convey information that can be completely supported by the provided document; this property has been referred in the literature to as faithfulness  \cite{rashkin-etal-2021-increasing,maynez2020faithfulness, dziri-etal-2021-neural,durmus-etal-2020-feqa} and attribution \cite{rashkin2021measuring}.  In our annotation set-up, we use similar definitions to the Attributable to Identifiable Source (AIS) framework of \citet{rashkin2021measuring}.  The full framework in that paper consists of a two-stage annotation process in which annotators first filter out responses that are deemed to be too vague or ill-formed to be evaluated for attribution. Since  \citet{rashkin2021measuring} found that more than 90$\%$ of the conversational responses in their study were interpretable, we have our annotators focus solely on attribution.

\paragraph{Not Attributable} These are responses that contain at least some information that cannot be verified given the evidence, regardless of whether that information is factually true in the real world.  This includes statements that are  relevant but not fully supported by the background information (hallucinations), statements that explicitly contradict the background information, and off-topic responses about information completely external to the evidence sources. In a pilot study we attempted to separate these three subcategories, but the boundaries between them turned out to be difficult to define and annotate.

\paragraph{Generic} Responses that fall into this category are general enough to fit into a large number of possible contexts  \cite{li2016diversity}. Examples include ``I don't know about that'' and ``Hello there!''. Even when the responses are ostensibly about the same topic as the document, they are vague and do not provide new information. 
Nevertheless, such responses may be useful for various conversational purposes: back-channeling, expressing uncertainty, or diverting the conversation from ambiguous or controversial topics.

\subsection{Collecting Prompt-Query-Reply Triples}
As described in Section~\ref{sec:setup}, we collect data using outputs from four models---T5, GPT2, DoHA, and \CTRL{}. We train a version of each model on each of the three datasets (\textsc{WoW}, \textsc{TopicalChat} and \textsc{CMU-DoG}) and generate responses using the test portion of the dataset. For more details on training and hyperparameters, refer to Appendix~\ref{sec:hyperparam}. We select at least 1000 examples from each dataset-model pair.
We filter and remove toxic responses using the Google Perspective API. This yields 12288 examples in total.
\begin{figure*}
\centering
  
    \subfigure[\label{fig:cmu_begin}\footnotesize Breakdown by model class.]{\includegraphics[width=0.48\linewidth]{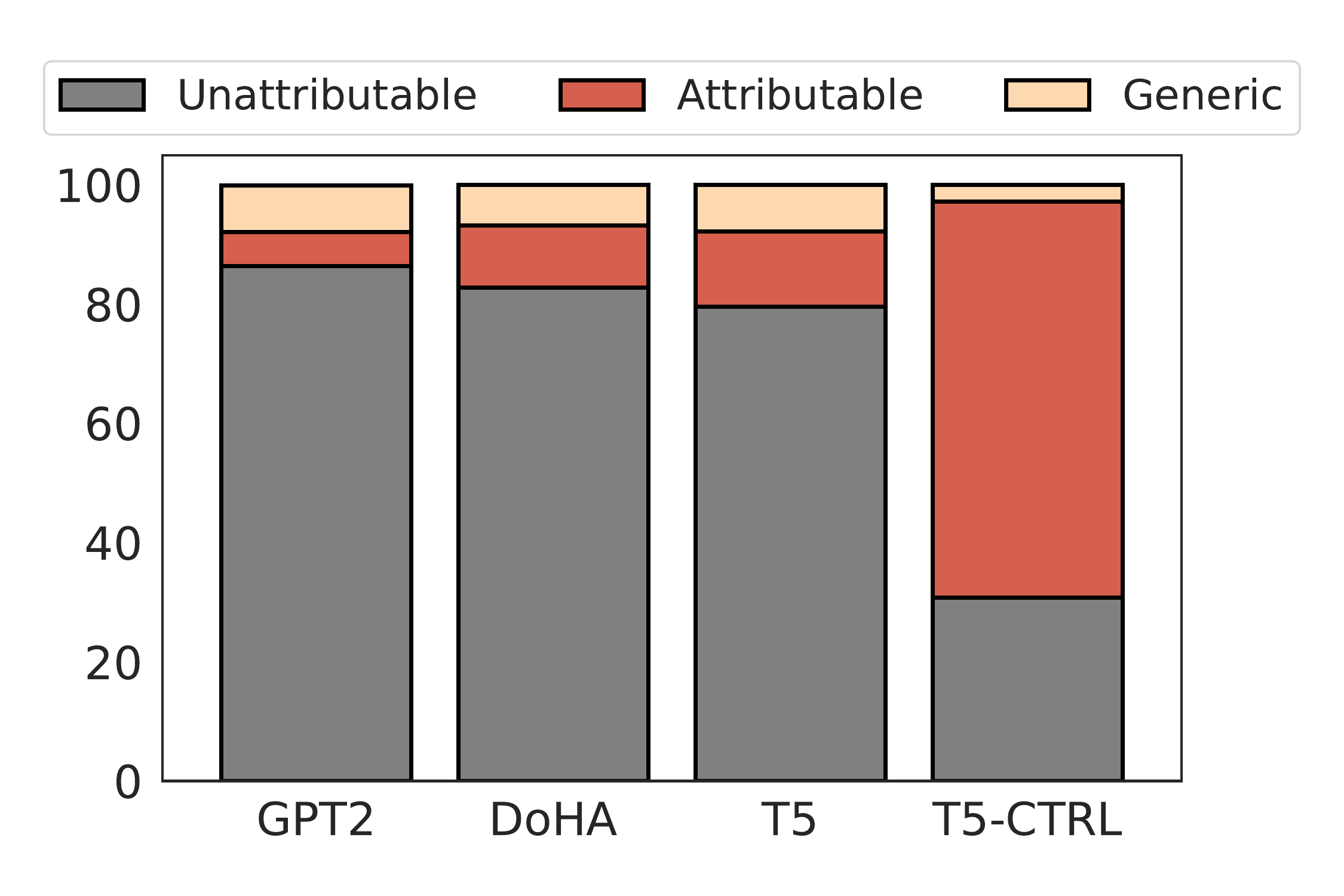}}
       \subfigure[\label{fig:wizard_begin}\footnotesize Breakdown by dialogue corpus used for training.]{\includegraphics[width=0.48\linewidth]{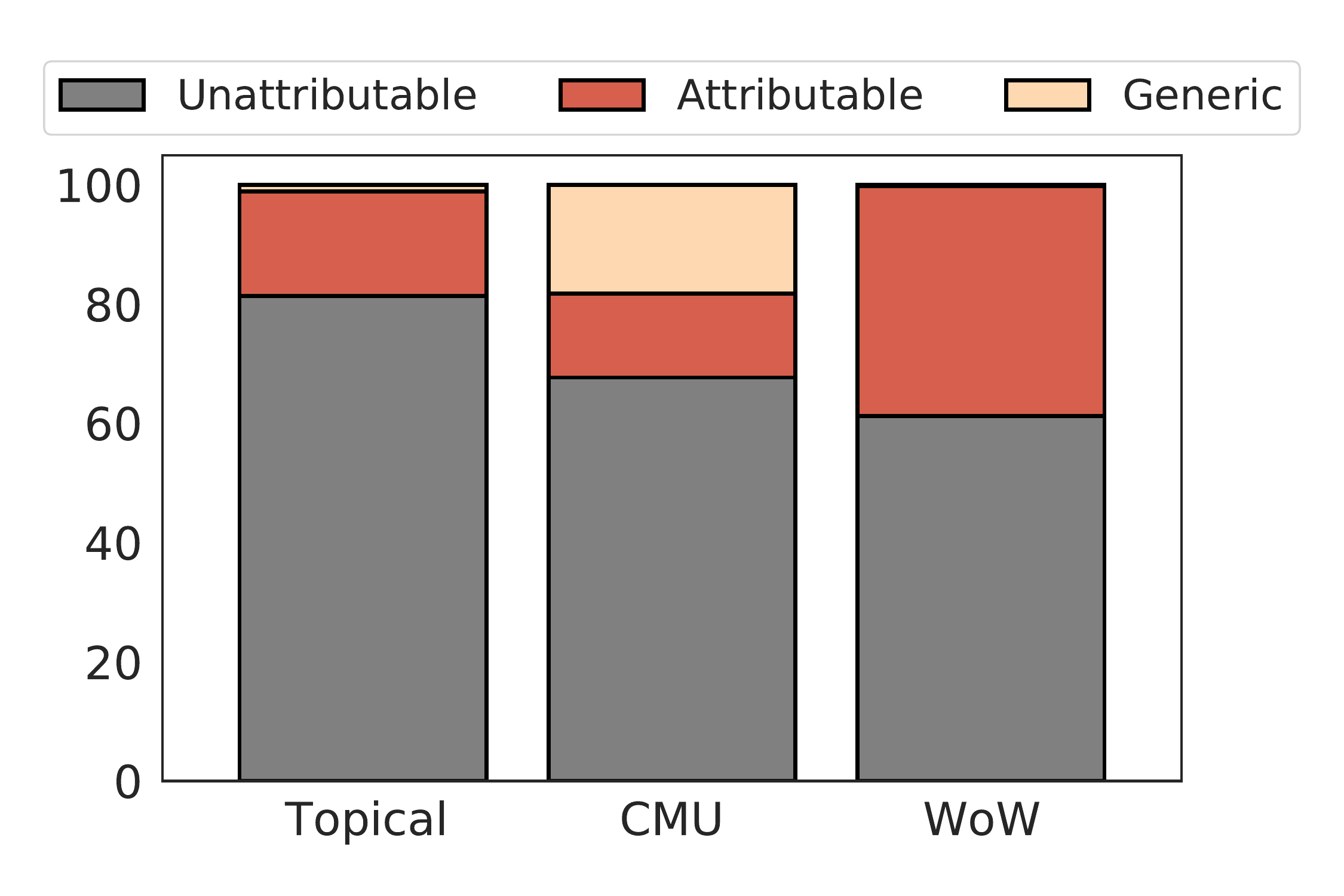}}
    \caption{\small Breakdown of \begindata{} response categories across models (left) and training corpora (right).}
    \label{fig:breakdown_gold}
\end{figure*} 
\subsection{Annotating  Prompt-Query-Reply Triples}
We present annotators with a knowledge snippet $\mathcal{K}$, the previous turn $u_{n-1}$ and a generated response $\bar{u}_{n}$, and ask them to select which of the three categories fits $\bar{u}_{n}$ best. 
For the exact annotation instructions, see Appendix~\ref{sec:annotationprotocol}.
To obtain high quality data, we assign three annotators to each example and report results based on majority vote. We exclude examples where each of the three annotators assigned a different category, making it impossible to compute a majority vote.

\paragraph{Annotation Quality}
 To ensure that the annotators understood the task, we use the following manual quality control procedure. In the first stage, we train the annotators by running two pilot annotation batches ($\sim100$~examples each).  After each batch, we manually grade the answers for compliance with instructions, and provide feedback explaining any misconceptions.  After the training stage, we launch the main annotation round for the full set of 12k examples.  During this round, we intermittently check responses after every 3k completed annotations to examine the annotation quality. %
 This procedure resulted in high inter-annotator agreement (a Krippendorff's alpha of 0.7).

\subsection{Dataset Analysis}
\label{sec4:data_analysis}
\BEGIN{} is intended as a test benchmark; as such, it does not have a training portion: We only create development ($10\%$) and test ($90\%$) partitions. 
We include examples from \BEGIN{} in Table~\ref{tab:benchmark:wow_cmu_begin_dist} along with the label breakdown. Overall, the models generated a substantial number of unattributable responses ($70\%$). As
Figure~\ref{fig:cmu_begin} shows, this proportion was higher for \textsc{GPT2}, \textsc{DoHA}, and \textsc{T5}, whereas \CTRL{} generated the lowest proportion of unattributable responses ($30.8\%$).  This indicates that \CTRL{}, which is explicitly designed to discourage unattributable responses, is moderately successful at its goal. Figure~\ref{fig:wizard_begin}, which breaks the results down by training corpus, shows that models trained on \textsc{TopicalChat} produce the highest amount of unattributable responses followed by \textsc{CMU-DoG} and \textsc{WoW}. 
This is consistent with recent analyses on \textsc{WoW}, \textsc{CMU-DoG} and \textsc{TopicalChat} which revealed that more than 60\% of the ground-truth  responses are unattributable to the knowledge \cite{dziri2022origin,rashkin2021measuring}.

\subsection{The Need to Measure Attribution}
Our analysis of the responses produced by the systems we trained highlights the potential pitfalls of language-model-based dialogue systems, especially when deployed in real-world scenarios across a broad range of domains where hallucinations pertaining to vital information may produce undesirable user experiences---e.g., healthcare \cite{laranjo2018conversational, jovanovic2020chatbots} and education  \cite{yang2019opportunities, kochmar2021automated}---and underscores the need for progress on both the modeling and the evaluation side.
Neural dialogue systems are optimized to mimic the distributional properties of the human-generated dialogue corpus used to train them. Because humans often include unattributable information in their utterances, language models trained on those corpora can replicate and perhaps even amplify the prevalence of unattributable responses at test time \cite{kang-hashimoto-2020-improved, dziri2022origin}. These findings call for robust evaluation metrics to uncover actionable insights about best practices of using such models and benchmarks.
We hope that \BEGIN{} will, as an evaluation benchmark, promote a strict standard for evaluation metrics, laying the ground for trustworthy dialogue systems.

\section{Evaluating Evaluation Metrics}
We next use \begindata \ to evaluate a range of evaluation metrics. In \S\ref{sec:metrics} we list the untrained metrics we use as well as metrics trained on existing resources, and in \S\ref{sec:adv} we describe a training set that we designed to train a classifier for the three response categories. We then describe the extent to which these metrics align with the \begindata{} categories and analyze the metrics' robustness. 

\subsection{Metrics}
\label{sec:metrics}

\paragraph{Lexical Overlap Metrics} This category includes $n$-gram-based metrics that compare the lexical similarity between the response $\bar{u}_{n}$ and the knowledge $\mathcal{K}$.\footnote{Note that we do not compare the generated responses to the gold responses as they may be unattributable (Sec \ref{sec4:data_analysis}).} We consider BLEU-4\footnote{\url{https://github.com/mjpost/sacrebleu}} \cite{papineni2002bleu}, ROUGE-L\footnote{\url{https://github.com/google-research/google-research/tree/master/rouge}} \cite{lin-2004-rouge}, and F1, which measures the word-level lexical overlap between $\bar{u}_{n}$ and  $\mathcal{K}$.

\paragraph{Semantic Similarity Metrics} These metrics compare the \textit{semantic} similarity between $\bar{u}_{n}$ and $\mathcal{K}$. We consider BERTScore \cite{Zhang*2020BERTScore:}, which computes the similarity between $\bar{u}_{n}$ and $\mathcal{K}$ based on the cosine similarity of the sentence embeddings, as well as BARTScore \cite{NEURIPS2021_e4d2b6e6} %
and BLEURT \cite{sellam2020bleurt}; %
for implementation details, see Appendix~\ref{app:implementation-details-metrics}. 

\paragraph{Question-Based Metrics} We use Q$^2$ \cite{honovich-etal-2021-q2}, which computes a factuality score through asking and answering questions. Given a candidate response as input, Q$^2$ generates a corresponding question and identifies potential answer spans in the knowledge source $\mathcal{K}$ that can justify the question--answer pair \cite{durmus-etal-2020-feqa, wang2020asking}. It also computes an NLI-inspired similarity score between a candidate response and a predicted answer span in the knowledge source.

\paragraph{Inference-Based Metrics} 
Finally, we study the performance of NLI-based models, trained either on gold NLI benchmarks or on adversarially augmented silver data that we generate. We first describe the metrics trained on gold NLI datasets; we discuss our adversarially augmented dataset (\textsc{BEGIN-Adversarial}) in \S\ref{sec:adv}.
We use two transformer-based classifiers: T5-base \cite{raffel2020exploring} and RoBERTa-large \cite{liu2019roberta}.  We fine-tune them on MNLI \cite{williams-etal-2018-broad} and the dialogue inference dataset DNLI  \cite{welleck-etal-2019-dialogue}. For both datasets, we map the labels (entailment, contradiction, neutral) to the labels (attributable, unattributable, generic) in \begindata.

We also train classifiers on AugWow \cite{gupta-etal-2022-dialfact}, a synthetic dataset designed to evaluate factuality in dialogue systems. 
This dataset includes three categories: \textit{Supported} responses that are fully verified by $\mathcal{K}$, \textit{Refuted} responses that explicitly contradict $\mathcal{K}$, and responses with \textit{Not Enough Information} (NEI), which do not contain enough information to be verified or refuted by $\mathcal{K}$.  We map the labels (supported, refuted, NEI) to the labels (attributable, unattributable, generic) in \begindata.

\subsection{Adversarially Augmented Training Set}
\label{sec:adv}

\begin{figure*}[ht]
\centering
\includegraphics[width=2.1\columnwidth]{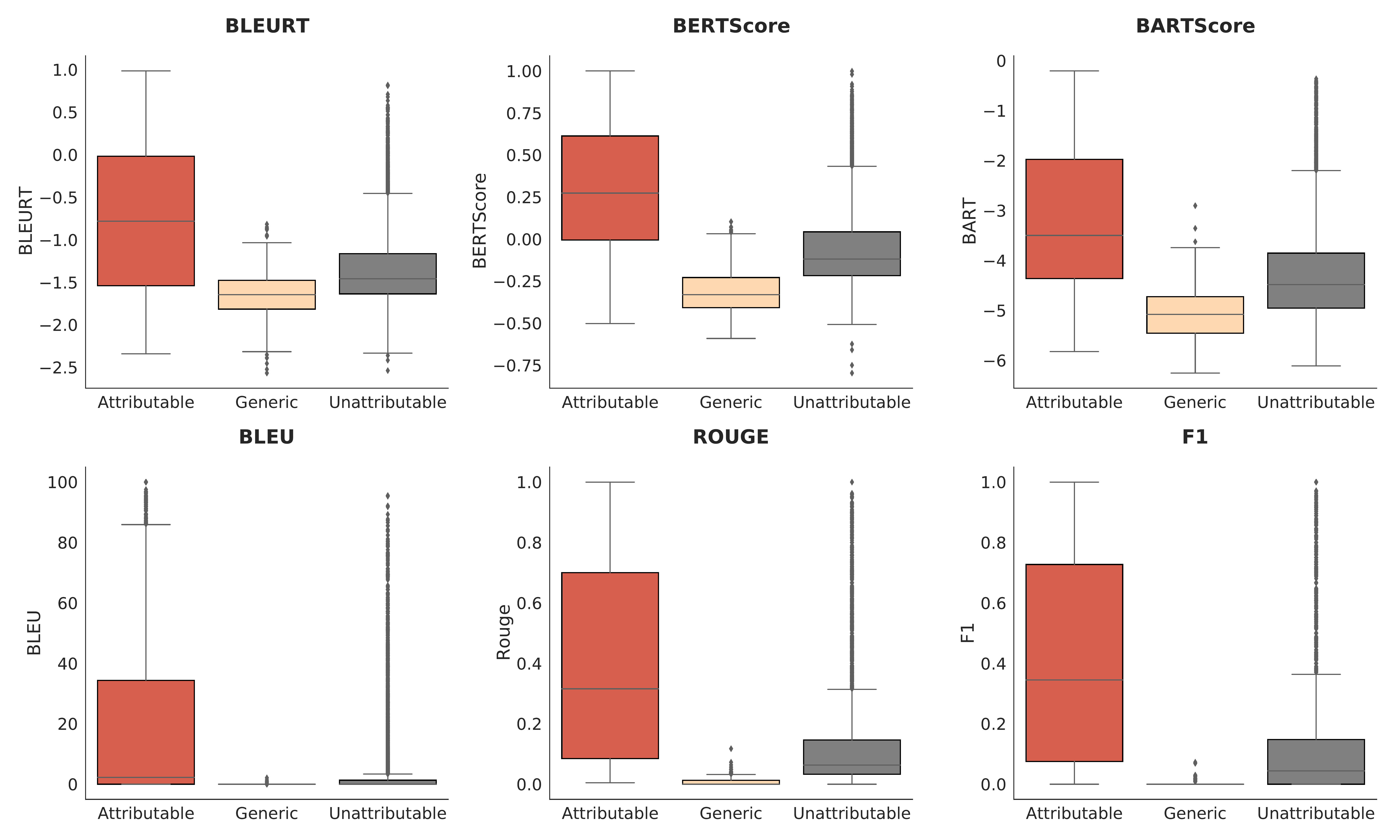}
\caption{\small The distribution of scores assigned by semantic similarity metrics (upper row) and lexical overlap scores metrics (lower row) to the \begindata{} test set. }
  \vspace{-15pt}
  \label{boxplot_lexical_semantic}
\end{figure*}

\begin{figure}[ht]
\centering
\includegraphics[width=0.85\columnwidth]{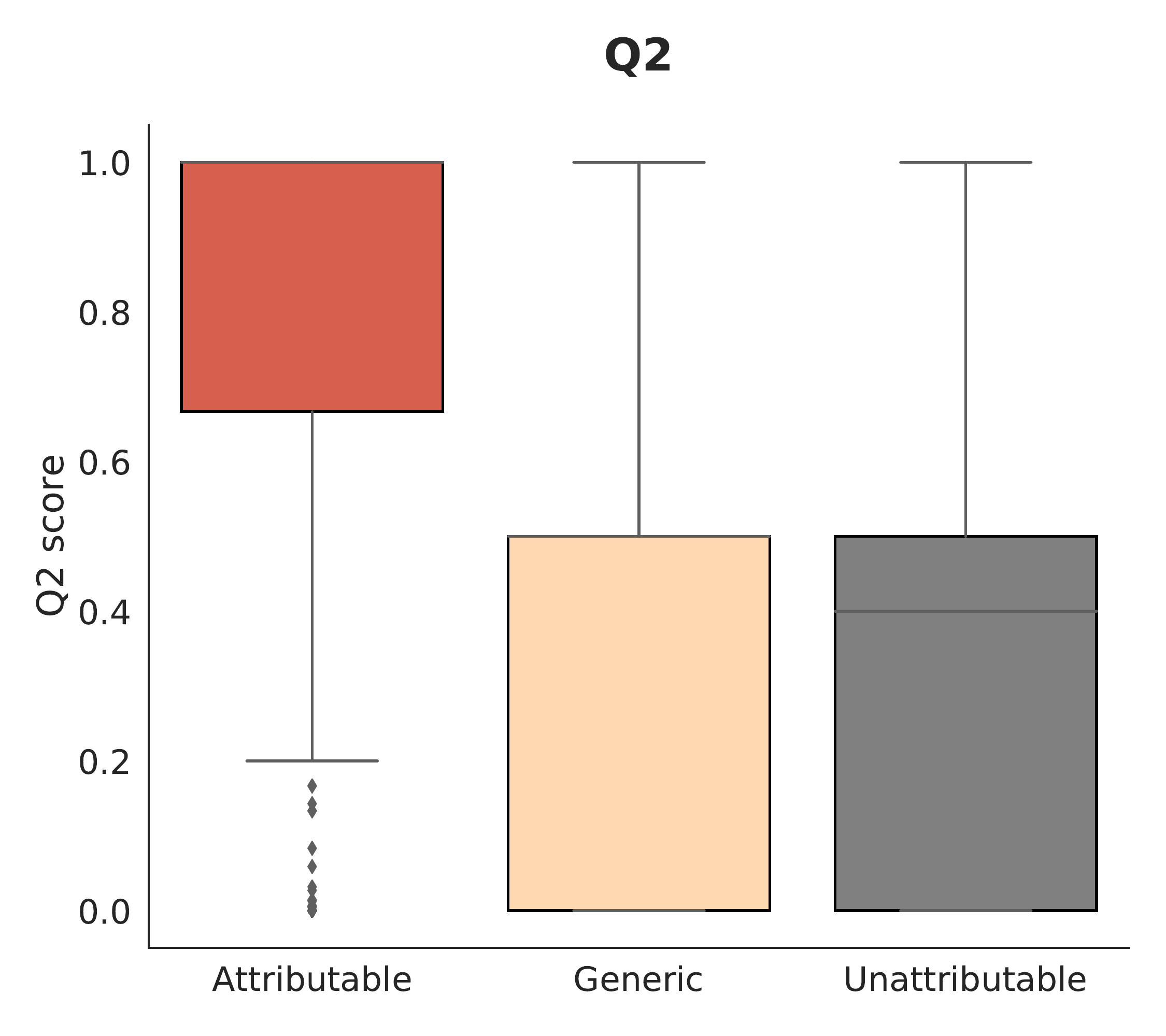}
\caption{\small The distribution of Q$^2$ scores for each of the three example categories in the \begindata{} test set. }
\label{q2_boxlot}
\end{figure}

This section describes our curated silver training set (\textsc{BEGIN-Adversarial}) for NLI-based attribution classifiers. This dataset includes 8k ($\mathcal{K}$, $\mathcal{H}$, $u_p$) triples that fit into the three categories: attributable, generic, and unattributable.

\paragraph{Attributable} Here we use the original human generated responses $u_g$ from \textsc{WoW}. To avoid human responses that contain opinions or generic chit-chat, we only use response that do not use first-person pronouns 
and where at least 25\% of the words in the response are contained in the evidence.

\paragraph{Unattributable}
To generate examples that are likely to be unattributable, but  are sufficiently challenging to distinguish from attributable ones as to be useful in training a classifier, we use multiple perturbation strategies.  First, we directly perturb the knowledge spans $\mathcal{K}$ from the \textsc{WoW} test set and then feed them to \textsc{GPT2} trained on \textsc{WoW}. We use three perturbation methods, each applied to a different  $\mathcal{K}$. First, we 
swap the subject and the object of  $\mathcal{K}$. Second, we replace up to two verbs with verbs of the same tense. Finally, we 
extract all mentioned entities from different dialogue examples using the SpaCy NER tagger \cite{honnibal2017spacy}, and replace up to two randomly chosen entities in the original $\mathcal{K}$ with entities of the same type. Manual inspection reveals that this usually results in responses that are hallucinations with respect to the original $\mathcal{K}$.

We also generate responses designed to specifically contradict $\mathcal{K}$, using two techniques. First, we directly negate the human response $u_g$ from \textsc{WoW} using the English Resource Grammar parser (ERG; \citealt{Fli:Ben:Oep:14}). 
Second, we replace adjectives in $u_g$ with their WordNet antonyms \cite{miller1998wordnet}. 

Lastly, we gather responses that are off-topic with respect to the information in the $\mathcal{K}$. For a given context, we randomly select a \textsc{WoW} gold response that was based on different  $\mathcal{K}$. To avoid easy-to-detect off-topic responses, we sample from conversations that were prompted by the same initial topic word as the target conversation.

\paragraph{Generic}
 Generic responses are generated from the \textsc{GPT2} model we trained on \textsc{WoW}, using a low softmax temperature of 0.4.

 \subsection{Results}
In this section, we report the performance of automatic metrics on the \begindata{} test set.

 \paragraph{Lexical and Semantic Metrics}
The distribution of scores is shown in Figure~\ref{boxplot_lexical_semantic}. For all metrics, the median score of fully attributable responses is higher than that of generic and  unattributable responses, as expected. In many individual cases, however, unattributable responses are scored quite highly, and there is some overlap in the distribution of scores across all three labels, particularly between generic and unattributable responses, indicating that it would be impossible to map these score ranges directly to the \begindata\ label taxonomy.
Higher scores do not always translate into more desirable response types: Even though a generic response would typically be preferable to an unattributable one in a knowledge-grounded dialogue system, the median scores are lower for generic responses than unattributable ones.

 \paragraph{Q$^2$}  Figure~\ref{q2_boxlot} shows a box plot for each \begindata{} class using the Q$^2$  metric. As in the case of the lexical and semantic metrics, Q$^2$ scores are typically higher for attributable responses but indistinguishable between generic and unattributable responses.  
 
\paragraph{Inference-Based Classifiers}

 Table \ref{tab:results} reports the performance of the NLI-based classifiers on \begindata{}. \textsc{BEGIN-Adversarial} substantially outperforms the classifiers trained on the gold datasets MNLI, DNLI and AugWoW even though it is a significantly smaller resource than those datasets. We also use MNLI as an intermediate fine-tuning dataset before fine-tuning on \textsc{BEGIN-Adversarial}.\footnote{We did not observe a similar improvement when using DNLI as an intermediate task.} 
We find that intermediate task fine-tuning can be beneficial when RoBERTa is used as the pretrained model ($\uparrow4.1$ on F1). 

Overall, our adversarially generated dataset provides better supervision for detecting our taxonomy than NLI-style datasets. 
This can be attributed to the fact that  NLI-style datasets are designed with a focus on detecting direct contradictions. 
By contrast, identifying unattributable responses requires detecting multiple types of unverifiable information including, but not limited to, contradictions. %
At the same time, none of the models exceed $46\%$ F1 score, showing that there is still room for improvement compared to human performance (over $95\%$ precision when comparing human annotations to the majority vote). Finally, \textsc{T5} and RoBERTa have similar F1 scores despite differences in model size and pretraining corpora, suggesting that simply scaling up the pretrained model may not be sufficient to make progress on this problem.

 \subsection{Are Metrics Measuring Attribution or Extractivity?} 
Do the metrics perform similarly on both challenging and easier examples? We adopt a density metric from \citet{grusky2018newsroom} to split the data into three groups---low, medium and high density---based on the extent to which they reuse language from the knowledge sources. Density represents the average length of the text spans in the responses that are copied from the knowledge. Extractive (high density) responses reuse the same phrases as the knowledge source, while abstractive (low density) responses may express the same meaning using a paraphrase. 

\begin{table}[ht]
    \scriptsize
    \centering
    \begin{tabular}{llccccc}
    \toprule
&\multicolumn{3}{c}{\bf{    \scriptsize Test set}}&	\multicolumn{3}{c}{\bf{Dev set}} \\
   \cmidrule(lr){2-4} \cmidrule(lr){5-7} 
\scriptsize{Finetuning data}& \bf P& \bf R & \bf F1	& \bf \bf P& \bf R& \bf  F1
\\\midrule
 \textbf{\textsc{T5}}  \vspace{0.15cm}\\

 MNLI		& 48.6 & 47.9	& 34.6 & 52.1 & 50.7 & 37.4 \\
DNLI  &	40.8 &	56.5 &	25.6  &  41.6 & 59.2 & 28.6 \\
 AugWow & 36.8 & 39.8 & 37.8  & 36.7 & 39.9	& 38.1  \\
 BEGIN-Adv. & 46.7 & 47.4 &{\bf 45.9} & 47.2 & 47.1 & {\bf46.3}\\
  { \; +MNLI} &	46.9 &	49.3 & { 45.3 }& 47.6& 49.4  & 46.1 \\
\midrule
\textbf{\textsc{RoBERTa}}  \vspace{0.15cm} \\
MNLI &  50.5 & 51.1	& 36.4 & 52.3&53.8 & 38.5\\
DNLI &	40.2 & 46.6 & 27.2&34.9 &  46.1 & 29.2\\
AugWow & 41.2 & 39.2 & 29.7 & 29.4  &	41.4&	29.1 \\
 BEGIN-Adv. & 42.6& {46.1} & 41.1& 49.2&	45.8&	41.1 \\
{ \; +MNLI}	& 44.8&	{45.9}&	{\bf45.2}	& 44.9	& 45.6	&{\bf45.1}\\
\midrule
\bf Human & 96.4&- & -&97.2&- & -\\
				\bottomrule		
    \end{tabular}
    \caption{Precision, recall and F1 of the classifier-based metrics created by fine-tuning T5 and \textsc{RoBERTa} on NLI datasets, AugWow and our adversarial training set. Scores are macro-averaged across labels on the \begindata{} test and dev sets.}
    \label{tab:results}
    \vspace{-5pt}
\end{table}

\begin{figure*}[ht]
\centering
\includegraphics[width=2\columnwidth]{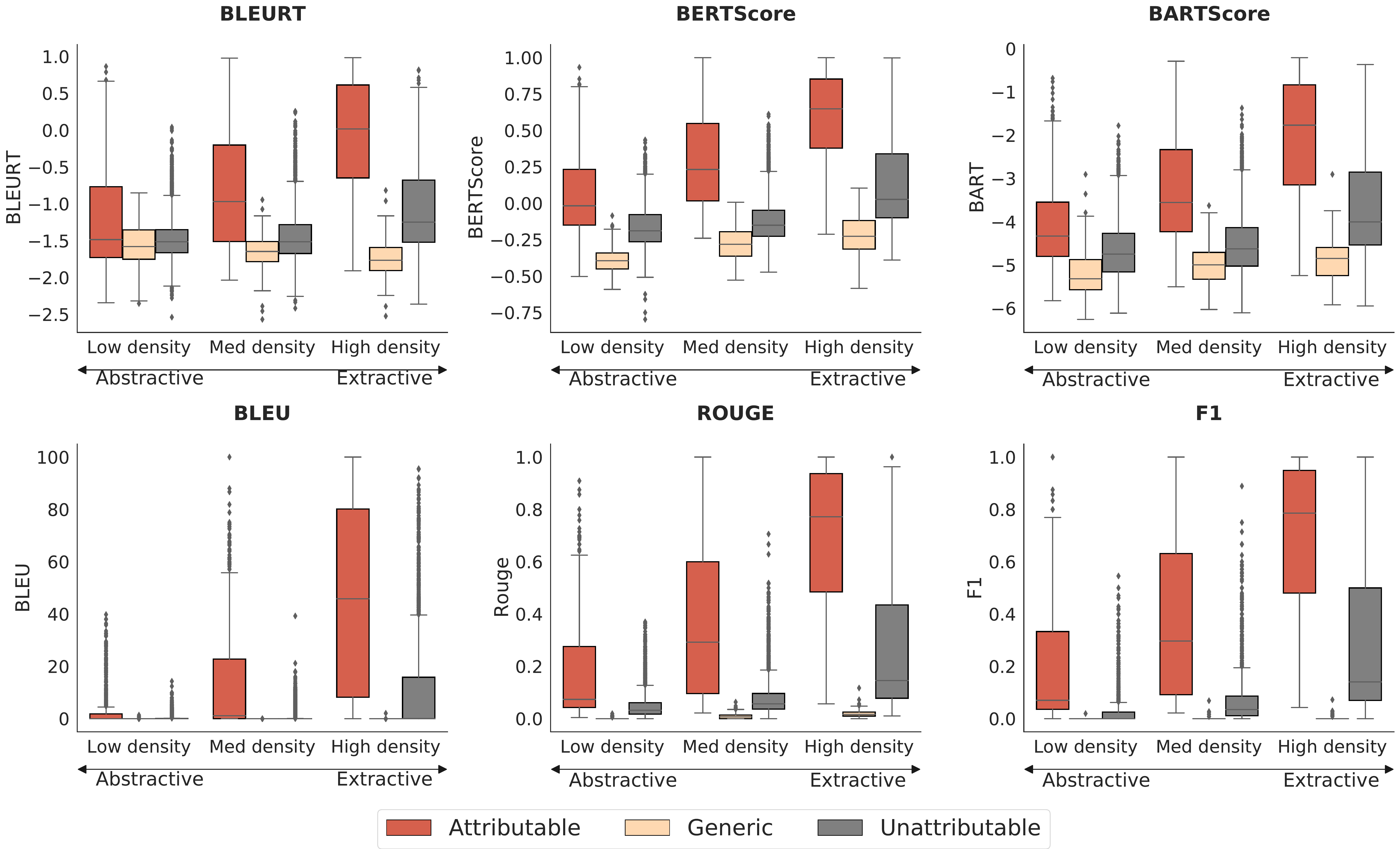}
\caption{\small Scores assigned to each of the three \begindata{} categories by semantic similarity metrics (upper row) and lexical overlap metrics (lower row), broken down by extractivity of the response (the extent to which it copies verbatim from the knowledge).}
\label{fig:metrics_density}
  \vspace{-15pt}
\end{figure*}

\begin{figure}[ht]
\centering
\includegraphics[width=1\columnwidth]{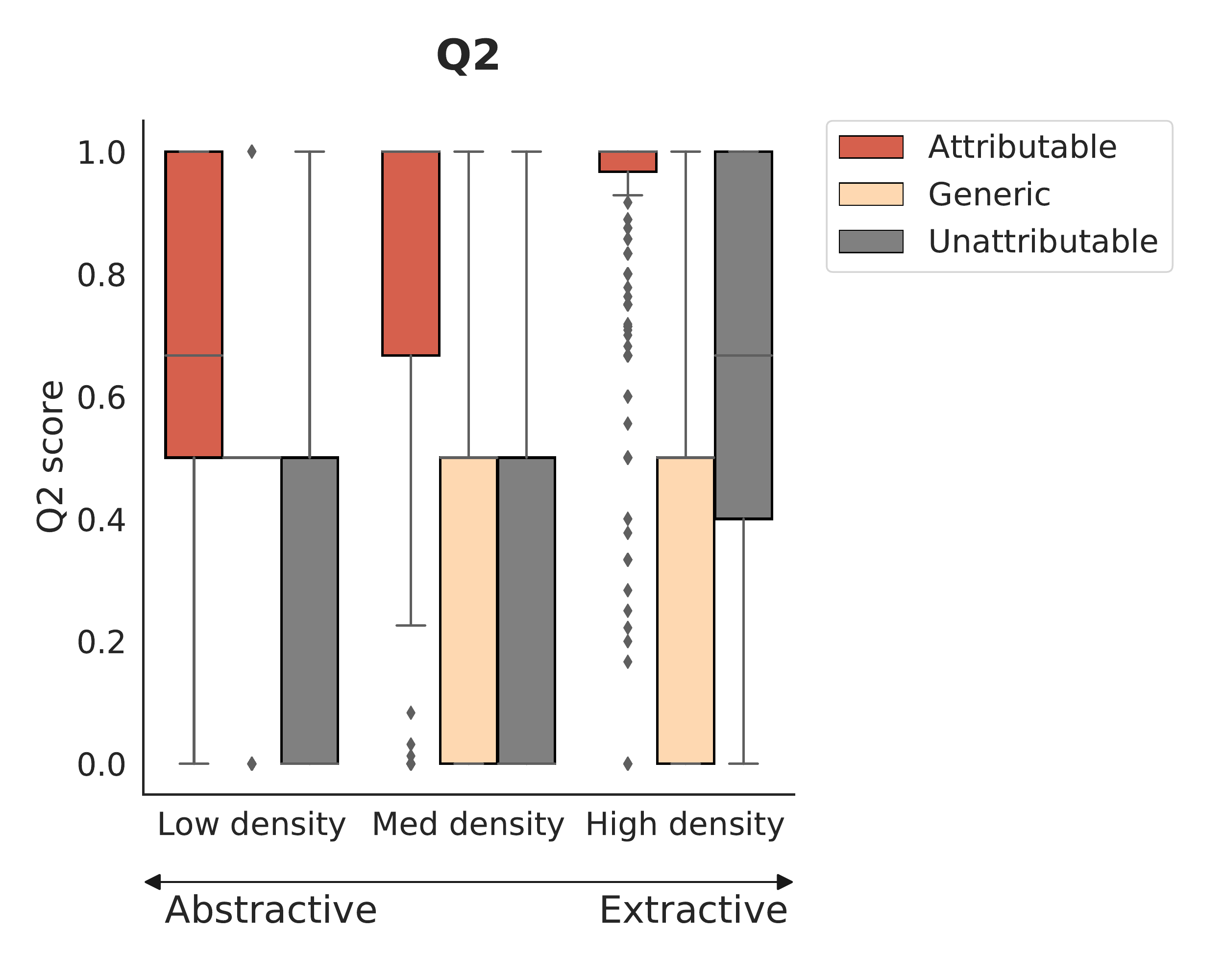}
 
\caption{\small Q$^2$ scores across extractive and abstractive responses on \begindata{} test. }
  \label{fig:q2_density}
\end{figure}

\paragraph{Results} Figures~\ref{fig:metrics_density} and~\ref{fig:q2_density} show the distributions across different levels of extractivity of the lexical and semantic metrics and the Q$^2$ score. 
We observe a common pattern across all metrics: high density responses for all categories (except \textit{generic} on BLEURT) score the highest, followed by medium density and low density responses.
The differences between the scores of the attributable, generic and unattributable categories are more pronounced in the more extractive responses, and less in the abstractive cases. Only Q$^2$, though generally unable to separate generic examples, maintains a clear separation between attributable and unattributable examples in the abstractive cases. Moreover, extractivity strongly influences the score assigned to attributable examples; an attributable response is likely to be scored much lower by all of these metrics if it is abstractive. Even more strikingly, unattributable extractive responses score higher on average than attributable abstractive responses in all metrics. 

\begin{figure*}[ht]
\centering
\includegraphics[width=2.1\columnwidth]{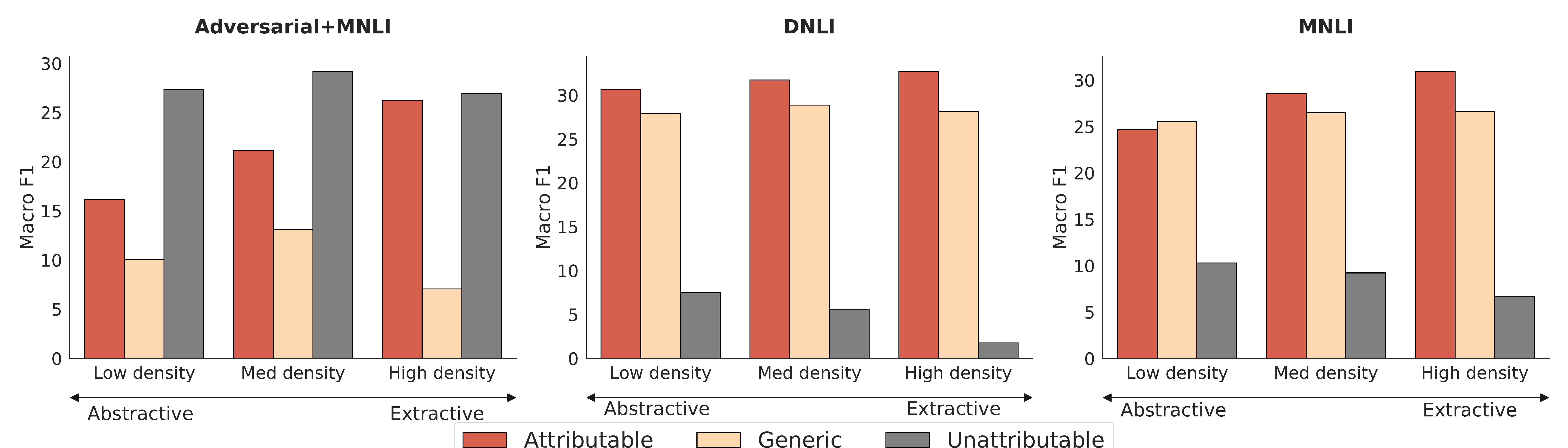}
\caption{\small Comparison of F1 scores of \textsc{RoBERTa}-based classifiers on \begindata{} categories with examples split by density (the extent to which the response copies verbatim from the knowledge). }
\label{fig:f1scoresbydensity}
\end{figure*}

We observe similar trends for the classifiers (Figure~\ref{fig:f1scoresbydensity}). The performance on  classifying attributable responses is much higher in extractive cases than in abstractive ones.  In contrast, the performance on unattributable responses is typically worse in the extractive cases. This pattern of results suggests that a response that is unattributable but has a high word overlap with the knowledge is very likely to be misclassified as attributable.  
In summary, we find that current metrics are relying on the spurious correlation between attribution and word overlap, and do not capture a deep understanding of the notion of attribution (cf. \citealt{mccoy-etal-2019-right}).

\subsection{Robustness to Distribution Shift} 
\begin{figure*}[ht]
\centering
\includegraphics[width=2.1\columnwidth]{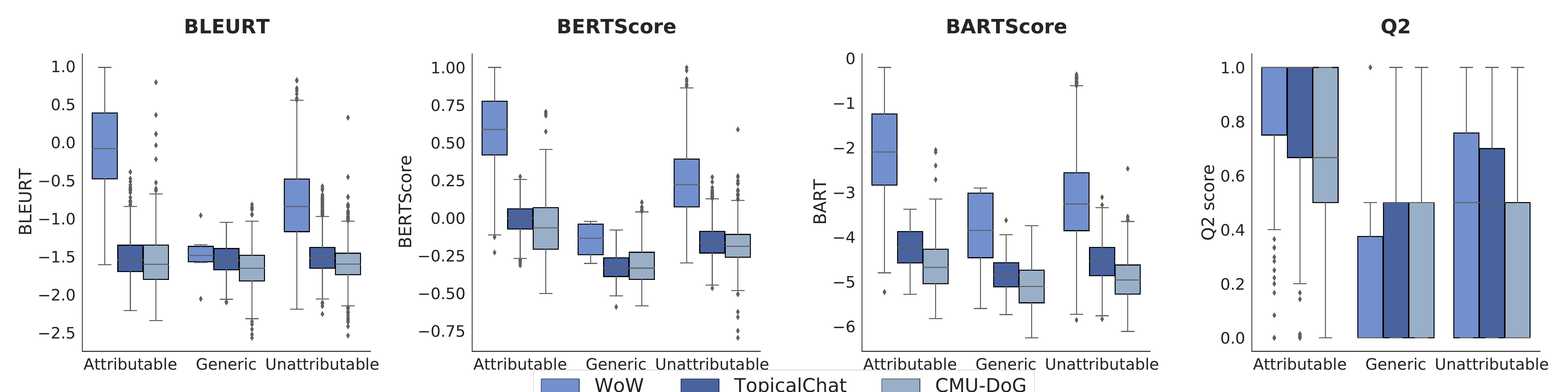}
\caption{\small Scores of the semantic and Q$^2$ metrics across the three dialogue corpora we used to train our models.}
\label{fig:boxplots_by_dataset}
\end{figure*}

\begin{figure*}[ht]
\centering
\includegraphics[width=2.1\columnwidth]{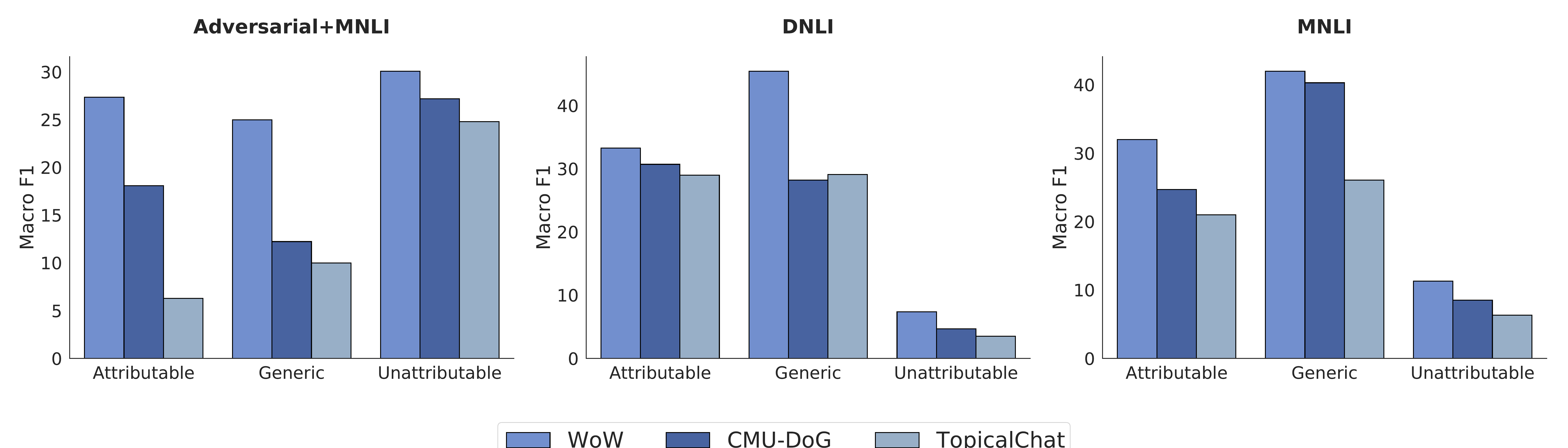}
\caption{\small Comparison of F1 scores of  RoBERTa classifiers on \begindata{} categories with examples split by benchmark. }
\label{fig:f1_by_dataset}
\end{figure*}

We further investigate the robustness of the metrics under distribution shift. Figure~\ref{fig:boxplots_by_dataset} shows the distributions of both semantic and Q$^2$ scores across the data broken down by source. All metrics\footnote{We observe similar results for lexical metrics.} rate responses from \textsc{WoW} in all categories significantly higher than responses derived from \textsc{CMU-DoG} and \textsc{TopicalChat}. Concerningly, attributable responses generated based on \textsc{CMU-DoG} and \textsc{TopicalChat} receive nearly identical scores to unattributable responses. Likewise, the F1 scores of all the classifiers (Figure~\ref{fig:f1_by_dataset}) are higher on the responses from \textsc{WoW} compared to the ones from \textsc{CMU-DoG} and \textsc{TopicalChat}. Specifically, classifiers tested on \textsc{TopicalChat} examples yield the worst F1 scores. For example, RoBERTA-MNLI's F1 score decreases by 10 points when tested on attributable responses from \textsc{TopicalChat} compared to \textsc{WoW}. 
In general, the metrics appear to perform poorly on datasets that have longer knowledge sources.  \textsc{TopicalChat} has on average  271 words in $\mathcal{K}$, followed by \textsc{CMU-DoG} and \textsc{WoW} which have 215 words, 27 words respectively. 
This shows that shorter knowledge spans correlates with higher metrics performance, pointing to the limited robustness of the metrics.

\section{Related Work}

\paragraph{Analysis of Evaluation Metrics in Natural Language Generation}
There is extensive interest in analyzing and meta-evaluating neural language generation (NLG) evaluation metrics \cite{gehrmann2022repairing, gehrmann2021gem}, for various tasks including machine translation \cite{freitag2021experts, mathur-etal-2020-tangled}, data-to-text generation \cite{dhingra2019handling}, summarization \cite{bhandari-etal-2020-evaluating,pagnoni-etal-2021-understanding,durmus-etal-2020-feqa, gabriel-etal-2021-go, fabbri2021summeval, durmus-etal-2022-spurious}, and dialogue generation \cite{yeh-etal-2021-comprehensive, durmus-etal-2022-spurious}.  Most of these studies have compared reference-free and reference-based evaluation metrics to human evaluation. 
For example, 
\citet{gabriel-etal-2021-go} measured the performance of automated metrics on summaries and compared certain dimensions such as sensitivity and high correlation with human scores. \citet{fabbri2021summeval}  analyzed metrics in summarization and released human-annotated data for faithfulness across 16 summarization models. We perform a similar meta-evaluation of existing automatic metrics in the context attribution in knowledge-grounded responses. Closest to our work is \citet{durmus-etal-2022-spurious}, who found that reference-free evaluation metrics of summarization and dialogue generation rely heavily on spurious correlations such as perplexity and length.

\paragraph{Metrics in Knowledge-Grounded Response Generation}
In contrast to the significant progress achieved in evaluating many NLG tasks, the evaluation of grounded response generation is a nascent research area \cite{shuster-etal-2021-retrieval-augmentation, rashkin2021measuring, dziri-etal-2021-neural}.  \citet{yeh-etal-2021-comprehensive} conducted a comprehensive study of existing dialog evaluation metrics. They measured properties such as engagingness and relevance but did not investigate the faithfulness of responses. While hallucination is well-studied in the context of summarization \cite{durmus-etal-2020-feqa,maynez2020faithfulness,nan-etal-2021,falke-etal-2019-ranking}, fewer researchers have looked into the problem of assessing hallucination in dialogue systems. 
\citet{dziri-etal-2021-neural} introduced a token-level critic that leverages a knowledge graph to identify hallucinated dialogue responses. 
\citet{rashkin2021measuring} proposed a human evaluation framework to assess output of dialogue models that pertains to the external world and utilized their evaluation framework for conversational QA tasks. \citet{dziri2022faithdial} introduced a faithful benchmark for information-seeking dialogues and demonstrated that it can serve as training signal for a hallucination critic, which discriminates whether an utterance is faithful or not.
An alternative approach for assessing faithfulness uses an auxiliary language understanding task, which measures whether a question answering system produces the same responses for the source document \cite{honovich-etal-2021-q2}. \BEGIN{} as a testing benchmark should be useful in developing similar metrics further.

 \paragraph{NLI and Adversarial Data for Grounded Dialogue Evaluation}
 In this work, we also investigate the performance of classifiers trained on NLI data, extending prior work that has proposed using NLI as a framework for evaluating conversational consistency \cite{welleck2018dialogue}. 
 \newcite{dziri2019evaluating} also used NLI to evaluate dialogue consistency. They generated a large-scale, noisy synthetic dataset of (premise, hypothesis) pairs tailored for dialogue, based on \newcite{Zhang2018Personalizing}.  
 We also explore training classifiers on adversarially augmented training data similar to concurrent work from  \newcite{gupta-etal-2022-dialfact} and \newcite{kryscinski2020evaluating}, which  proposed a synthetic dataset for determining whether a summary or response is consistent with the source document; this dataset was constructed by applying a number of syntactic transformations to reference documents (for a similar approach applied to NLI, see \citealt{min-etal-2020-syntactic}).

\section{Conclusion}
Contemporary knowledge-based dialogue systems that rely on language models often generate responses that are not attributable to the background knowledge they are expected to convey. We present \textsc{Begin}, a new benchmark to advance research toward robust metrics that can assess this issue. 
We use \textsc{Begin} to comprehensively evaluate a broad set of existing automatic metrics. We show that these metrics rely substantially on word overlap and fail to properly rank abstractive attributable responses as well as generic responses. They also struggle under distribution shift, assigning low scores to attributable responses grounded on long knowledge sources. 
We hope that this work will spur future research on building robust evaluation metrics for grounded dialogue systems.

\section*{Acknowledgements}
We are grateful to the anonymous
reviewers for helpful comments.
We thank Dipanjan Das, Vitaly Nikolaev, Sebastian Gehrmann,  Roee Aharoni, Jennimaria Palomaki,  Tom Kwiatkowski, Michael Collins and Slav Petrov for helpful discussions and feedback. We also thank Ashwin Kakarla and his team for
helping with the annotations.
\appendix

{

}

\section{\textsc{Begin} Annotation Protocol} 
\label{sec:annotationprotocol}
Each worker was given a document, previous turn in a conversation and a generated response (either by \textsc{T5}, \textsc{GPT2}, \textsc{DoHA} or \CTRL{}).  They were asked to evaluate the response as either fully attributable, not attributable, or too generic to be informative. They also were provided with multiple examples with explanations for each category. The exact instructions were as follows:
\begin{mdframed}[leftmargin=0pt,rightmargin=0pt]
\small
Which of these best describes the highlighted utterance?
\begin{itemize}
    \item[$\circ$] {Generic: This utterance is uninformative (too bland or not specific enough to be sharing any new information) }
    \item[$\circ$] {Contains \emph{any} unsupported Information:
This utterance is sharing information that cannot be fully verified by the document.  It may include  false information, unverifiable information, and personal stories/opinions.}
    \item[$\circ$] {\emph{All} information is \emph{fully} supported by the document: This utterance contains only information that is fully supported by the document.}
\end{itemize}
\end{mdframed}

\section{Implementations}
\label{sec:hyperparam}

\paragraph{GPT2, T5} We implement these models using the TensorFlow Huggingface Transformers library \cite{wolf-etal-2020-transformers}. During training, we use the Adam optimizer \cite{DBLP:journals/corr/KingmaB14} with Dropout \cite{srivastava2014dropout} on a batch size of $32$ with a learning rate of $6.25 \times 10^{-5}$ that is linearly decayed. The maximum dialogue history length is set to $3$ utterances. The model early-stops at epoch \{6, 10, 10\} respectively for \textsc{WoW}, \textsc{CMU-DoG} and \textsc{TopicalChat}.

\paragraph{\CTRL{}} We reproduce the results from \cite{rashkin-etal-2021-increasing}, following the training details in that paper.

\paragraph{DoHA} We use the code and the pre-trained model on \textsc{CMU-DoG} that are publicly available by the authors at their Github's account \footnote{\url{https://bit.ly/3bBup2M}}. For \textsc{WoW} and \textsc{TopicalChat}, we follow closely the authors' training procedure described in \citep{prabhumoye-etal-2021-focused} and we train two models on both datasets.

For each dataset, we save the best model based on the validation set. We use nucleus sampling with $p=0.9$.

\section{Model-Based Metrics}
\label{app:implementation-details-metrics}

\paragraph{Semantic Similarity Models}
We use BERTScore version 0.3.11. with the
DeBERTa-xl-MNLI model \cite{he2020deberta}, which is the recommended model as of the time of investigation. For BLEURT, We use the recommended
BLEURT-20 checkpoint \cite{pu-etal-2021-learning}. For BARTScore, we use the latest publicly available checkpoint (accessed March 2022) from \url{https://github.com/neulab/BARTScore}.

\clearpage

\bibliography{grounded_dialog_metric,anthology}
\bibliographystyle{acl_natbib}

\end{document}